# AN INVESTIGATION OF BENFORD'S LAW DIVERGENCE AND MACHINE LEARNING TECHNIQUES FOR INTRA-CLASS SEPARABILITY OF FINGERPRINT IMAGES

Aamo Iorliam*[1], Orgem Emmanuel[2], and Yahaya I. Shehu[3]
[1,2] Department of Mathematics & Computer Science, BSU, Makurdi, Nigeria
[3] Shehu Shagari College of Education, Sokoto, Nigeria
Corresponding author: Iorliam, A. (aamoiorliam@gmail.com).

## ABSTRACT

*Protecting a biometric fingerprint database against attackers is very vital in order to protect against false acceptance rate or false rejection rate. A key property in distinguishing biometric fingerprint images is by exploiting the characteristics of these different types of fingerprint images. The aim of this paper is to perform an intra-class classification of fingerprint images using the Benford's law divergence values and machine learning techniques. The usage of these Benford's law divergence values as features fed into the machine learning techniques has proved to be very effective and efficient in the intra-class classification of biometric fingerprint images. The effectiveness of our proposed methodology was demonstrated on five datasets, achieving very high intra-class classification accuracies of 100% for the Decision Tree and Convolutional Neural Networks (CNN). However, the Naïve Bayes, and Logistic Regression achieved accuracies of 95.95%, 90.54%, respectively. These results showed that Benford's law features and machine learning techniques especially Decision Tree and CNN can be effectively applied for the intra-class classification of fingerprint images.*

**Keywords:** Benford's law, divergence values, machine learning techniques, intra-class separability, fingerprint images.

# 1. INTRODUCTION

Biometric experts have been dependent on fingerprints over the years for verification and identification purposes. There exist different types of fingerprint images which include contact-less acquired fingerprints, optically acquired fingerprints, and synthetically generated fingerprints [1], [2]. Since these fingerprints are used for different purposes, they should not be intentionally or unintentionally used for another purpose as this may cause a serious security threat [3]. Therefore, this paper performs an investigation of machine learning techniques for intra-class classification of biometric images to classify biometric images of different types that have the same modality.

It has been reported in the literature that since 1938, Benford's law [4] has proved beyond reasonable doubt that it possesses the capability to detect/classify original/untampered data from fake/tampered data. This interesting law (Benford's law) is therefore adapted for the intra-class separability of biometric images. Firstly, the gray-scale fingerprint images are used to calculate the first digit distribution of the JPEG coefficients using Equation 2. Furthermore, Equation 3 is used to calculate Benford's law divergence values. These divergence values are fed as inputs into the machine learning techniques such as the Naïve Bayes, Decision Tree, Logistic Regression, and Convolutional Neural Networks.

The applicability of this research is that it can serve as a preliminary forensic tool in classifying different types of fingerprint images for forensics and biometric applications. The major contributions of this work are summarized as follows:

i. We propose a novel use of Benford's law divergence values to improve the intra-class classification of fingerprint images.
ii. We provide a detailed analysis of the novel intra-class classification of fingerprint images using an empirical study based on theoretical and empirical perspectives.
iii. We used only six Benford's law features (reduced features) and performed intra-class classification with a high-performance evaluation.

The rest of the paper is organized as follows. Related works are described in Section 2. Section 3 describes our experiments, datasets used, divergence metric determination, data pre-processing for intra-class separability of fingerprint images. Results and discussions are presented in Section 4. Conclusion and future work are presented in Section 5.

## 2. RELATED WORKS

Benford's law has proved to be very effective in detecting forged/tampered images. The Benford's law was first discovered in 1881 by Simon Newcomb, where he noticed that the first pages of the logarithm table containing the first digits were worn more than the last pages of the logarithm table, which meant that people were looking up for numbers starting with 1 more often than numbers starting with 2, and so on [5]. Unfortunately, Newcomb could not prove why the theory and formula worked. Then in 1938, Frank Benford proposed the Benford's law, also referred to as the first digit law, which states that multi-digit numbers beginning with 1, 2, or 3 appear more frequently than multi-digit numbers beginning with 4, 5, 6, 7, 8 and 9 [3],[4].

Benford's law maintains that the digit 1 will be 30.1% of the time the leading digit when considering a genuine data set of numbers, the digit 2 will be 17.6% of the time the leading digit, and each subsequent digit, 3 through 9, will be the leading digit with decreasing frequency in an expected occurrence as illustrated in Figure 1.

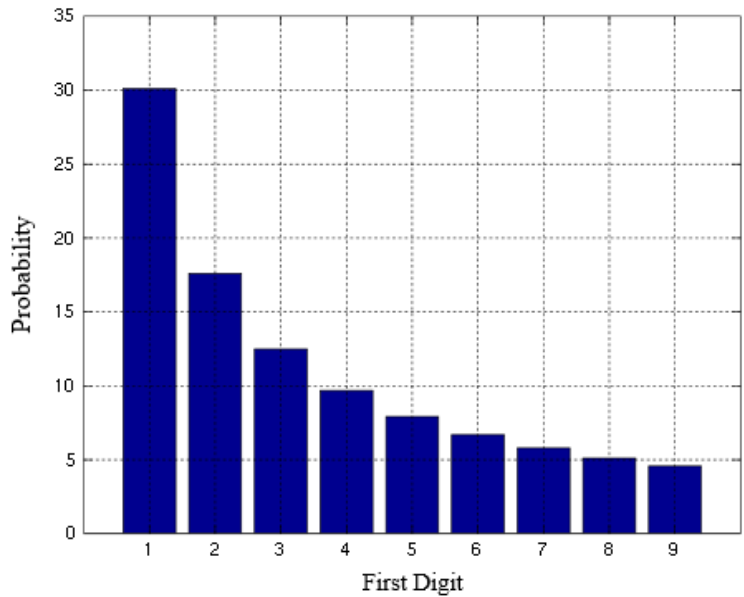

**Figure 1**: The first digit probability distribution of Benford's law [3], [4].

Taking into consideration the Most Significant Digit, where 0 is not included, and a dataset under investigation satisfies Benford's law, the standard Benford's law is expressed as follows:

$$p(x) = \log_{10}(1 + \frac{1}{x}), x = 1, 2, 3, \ldots, 9 \qquad (1)$$

where *x* is the first digit of the number and *p(x)* refers to the probability distribution of *x*.

Since the inception of the Benford's law, it is expected that naturally generated datasets should obey this law, whereas tampered or randomly generated datasets should deviate from this law. This inherent characteristic of the Benford's law can lead to important applications in forensics such as detecting anomalies or fraud in a given dataset [6], [7] or classifying different types of biometric images [3], [8].

Intra-class separability of biometric images means the classification of biometric images that appear to be identical (closely related) [6]. For instance, fingerprint images such as contact-less acquired latent fingerprints, optically acquired fingerprints and synthetically generated fingerprints are closely related due to the fact that they cannot be easily classified based on their physical appearance, hence classifying them could be referred to as intra-class classification of fingerprint images [3]. Even though, intra-class classification of biometric images seem to be a novel area, Table 1 summarises the related work in this area.

**Table 1**: Summary of Related Works

| Research Title/Author | Implementation Strategy | Advantages |
|---|---|---|
| Benford's law based detection of latent fingerprint forgeries on the example of artificial sweat printed fingerprints captured by confocal laser scanning microscopes [9] | The model employed the use of Benford's law and WEKA's Bagging classifier in 10-fold stratified cross-validation. | Differentiated between real latent fingerprints and printed fingerprints using Benford's law in the spatial domain. |
| Using Benford's law divergence and neural networks for classification | Applied Benford's law and neural networks for the | A novel approach for the source identification of captured biometric images. |

| Research Title/Author | Implementation Strategy | Advantages |
|---|---|---|
| and source identification of biometric images [3] | classification of biometric images. | |
| On the use of Benford's law to detect JPEG biometric data tampering [6] | Used Benford's law with SVM in the biometric fingerprint tampering detection and separability of fingerprint images. | A novel approach to fight against insider attackers and hackers for securing biometric fingerprint images. |
| On digitized forensics- novel acquisition and analysis techniques for latent fingerprints based on signal processing and pattern recognition [10] | A thesis that contributed to digital forensics, latent fingerprint processing, and latent fingerprint forgery detection. | A novel contribution to digitized forensic and latent fingerprints. |
| On the use of Benford's law to detect GAN-generated images [11] | Used Benford's law features with a simple Random Forest classifier. | Discriminated GAN-generated images from natural photographs. |
| An investigation of machine learning techniques for intra-class separability of biometrics images (Our Proposed Method) | Used Benford's law features with Naive Bayes, Decision Tree, Logistic Regression, and CNN. | Effectively reduce features and achieved high intra-class separability of fingerprint images. |

To the best of the researcher's knowledge and review presented, this paper presents for the first time the novel use of Benford's law features with machine learning techniques to accurately classify the fingerprint images.

## 3. EXPERIMENTAL SETUP

The goal of this experiment is to utilize the acquired Benford's law divergence values from fingerprint images as proposed by Fu *et al*., [12] and Iorliam *et al*., [6] for the separability of biometric images.

In essence, the intra-class classification of biometric fingerprint images is performed on Benford's law divergence values of DB1, DB2, DB3, DB4, and the artificially acquired contact-less latent fingerprints images. The extracted fingerprint features are fed into the Naive Bayes, Decision Tree, Logistic Regression, and Convolutional Neural Networks (CNN) algorithms as input data for separability purposes. The need for the separability of fingerprint images is discussed in Section 3.1.

### 3.1. Need for Intra-Class Separability of Fingerprint Images

The two key uses of biometrics are verification and identification. Verification is usually a 1-to-1 matching, whereas identification is a 1-to-many matching. For more than a century, fingerprints have been used for identification purposes [6], [13]. Fingerprints are used for different purposes as explained by Iorliam *et al.* [6]. Therefore, it is very important to avoid using a particular type of fingerprint for another purpose either intentionally or unintentionally. This can be achieved by studying the characteristic of the type of the different fingerprints and as such identifying the source of the captured fingerprint images. This could be possible if the source hardware that captured the fingerprint image is identified [14]. One way to do this is by utilizing Benford's law divergence values with machine learning techniques to achieve the fingerprint images separability.

### 3.2 Data Sets Used

The FVC2000 fingerprint dataset which consists of four different fingerprint databases (DB1, DB2, DB3, and DB4) is used in this research. Furthermore, artificially printed contact-less acquired latent fingerprint images are used for this research. Therefore, a total of five different datasets are used for testing our proposed model. The first four (4) set of datasets each contains 80 greyscale fingerprint images. The datasets are accessible at: http://bias.csr.unibo.it/fvc2000/databases.asp [15]. While the artificially printed contact-less acquired latent fingerprint images have 48 fingerprint biometric images and are available at: https://link.springer.com/chapter/10.1007%2F978-3-642-40779-6_19 [1]. Other details about the datasets used are provided in Table 2.

Table 2: Summary Description of Datasets

| Source | Dataset | Sensor Type | Sample No |
|---|---|---|---|
| FVC2000 [15] | DB1 | Low-cost Optical Sensor captured by "Secure Desktop Scanner". | 80 |
| | DB2 | Low-cost Optical Capacitive Sensor captured by "TouchChip" | 80 |
| | DB3 | Optical Sensor "DF-90" | 80 |
| | DB4 | Synthetically generated images from Synthetic Generator | 80 |
| Hildebrandt et al., [1] | DB5 | Artificially acquired contact-less acquired fingerprint images | 48 |

**3.3 Divergence Metric Determination**

Benford's law divergence values are obtained based on the biometric fingerprint dataset used in this paper described in Section 3.2. The divergence metric is used to show how close or far a particular dataset is, using the standard or generalised Benford's law. In any case, smaller divergence yields a better fitting. In this paper, the first digit distributions of the JPEG coefficients are extracted from the gray-scale images as demonstrated by Iorliam, *et al.*, [3].

Fu *et al.*, [12], extended the standard Benford's law to the Generalised Benford's law which closely follows the logarithmic law as expressed in Eq. (2).

$$p(x) = N \log_{10}(1 + \frac{1}{s + x^q}) \qquad (2)$$

where N is the normalization factor which makes *p(x)* a probability distribution. The model parameters s and q describe the distributions for different fingerprint images and different compressions of the Quality Factor (QF). Through experiments, Fu, *et al.*, [12], provided values for N, s, and q using the Matlab toolbox, which returns the Sum of Squares due to Error (SSE). The N, s, and q values are as shown in Table 3 for the Generalised Benford's law experiments.

**Table 3**: Model parameters used for the generalised Benford's law [12]

| Q-factor | Model Parameters | | | Goodness-of fit (SSE) |
|---|---|---|---|---|
| | N | q | s | |
| 100 | 1.456 | 1.47 | 0.0372 | $7.104e - 06$ |
| 90 | 1.255 | 1.563 | −0.3784 | $5.255e - 07$ |
| 80 | 1.324 | 1.653 | −0.3739 | $3.06838e - 06$ |
| 70 | 1.412 | 1.732 | −0.337 | $5.36171e - 06$ |
| 60 | 1.501 | 1.813 | −0.3025 | $6.11167e - 06$ |
| 50 | 1.579 | 1.882 | −0.2725 | $6.05446e - 06$ |

To test for conformity of a particular dataset (fingerprint images) to Benford's law, one of the most common criteria used is the chi-square goodness-of-fit statistics test [3], [16], [17]. The chi-square divergence is expressed in Eq. (3).

$$x^2 = \sum_{x=1}^{9} \frac{(P'x - Px)^2}{Px} \quad (3)$$

where $P'x$ is the actual first digit probability of the JPEG coefficients of the fingerprint biometric images and $Px$ is the logarithmic law (Generalised Benford's law) as given in Eq. (2). In this study, the fingerprint datasets are singly compressed at a QF of 50 to 100 in a step of 10 [3]. The divergence is calculated as an average on all the datasets earlier described in Table 2.

**3.4 Data Pre-Processing for Intra-Class Separability of Fingerprint Images**

In this paper, the fingerprint datasets are transformed into a format that can be easily interpreted by the machine learning techniques under consideration. The available biometric fingerprint images are transformed and this results in 368 instances with 6 features and a Class Label as shown in Figure 2. The first 6 attributes are compression Quality Factors (QFs) ranging from 50 to 100 in a step of 10. The 7th attribute is the class label, which is represented as 0, 1, 2, 3, and 4 for DB1, DB2, DB3, DB4, and the artificially printed contact-less acquired latent fingerprint (contact-less), respectively. The pre-processing values are achieved with the help of Benford's law divergence which is explained in Sections 3.3.

|   | A | B | C | D | E | F | G |
|---|---|---|---|---|---|---|---|
| 1 | QF-50 | QF-60 | QF-70 | QF-80 | QF-90 | QF-100 | Class Label |
| 2 | 10.49719 | 10.80152 | 9.790624 | 8.71157 | 7.745091 | 7.761463 | 0 |
| 3 | 11.30665 | 10.93996 | 9.681446 | 8.675576 | 7.871395 | 8.270308 | 0 |
| 4 | 10.12913 | 10.14469 | 9.176165 | 8.658021 | 7.656819 | 7.184597 | 0 |
| 5 | 8.979308 | 9.382513 | 8.908421 | 8.30894 | 7.466352 | 6.91005 | 0 |
| 6 | 11.77114 | 11.00361 | 9.720688 | 8.637295 | 8.06052 | 8.575394 | 0 |
| 7 | 11.79356 | 11.31105 | 9.884672 | 8.865339 | 8.049554 | 8.608477 | 0 |
| 8 | 12.07603 | 11.50856 | 9.993754 | 8.806326 | 8.007224 | 8.698664 | 0 |
| 9 | 11.61334 | 11.30085 | 9.944549 | 8.701875 | 8.010162 | 8.350574 | 0 |
| 10 | 10.33387 | 10.20274 | 9.487949 | 8.407865 | 7.774554 | 7.828174 | 0 |
| 11 | 10.26303 | 9.98995 | 9.359262 | 8.556169 | 7.858129 | 8.139676 | 0 |
| 12 | 8.815993 | 9.117034 | 8.976531 | 8.549314 | 7.675995 | 7.306121 | 0 |
| 13 | 10.13061 | 9.755888 | 8.945211 | 8.413265 | 7.927866 | 8.046223 | 0 |
| 14 | 9.36388 | 9.357628 | 8.916166 | 8.358755 | 7.937299 | 7.41311 | 0 |
| 15 | 9.266528 | 9.366076 | 8.757959 | 8.609852 | 7.888462 | 7.700743 | 0 |
| 16 | 9.749225 | 9.79921 | 9.364893 | 8.716818 | 7.949279 | 7.712645 | 0 |
| 17 | 9.829509 | 9.552589 | 9.189725 | 8.738055 | 8.03372 | 7.990401 | 0 |
| 18 | 8.928493 | 8.876035 | 8.798206 | 8.396719 | 7.838692 | 7.221092 | 0 |
| 19 | 9.927483 | 9.600776 | 8.826528 | 8.501742 | 8.240397 | 8.034938 | 0 |
| 20 | 9.748196 | 9.231848 | 9.098562 | 8.715603 | 8.203311 | 7.513904 | 0 |
| 21 | 8.738349 | 8.561446 | 8.620333 | 8.544117 | 8.017608 | 7.300517 | 0 |
| 22 | 9.306565 | 8.922777 | 8.788508 | 8.374115 | 7.874848 | 7.872943 | 0 |
| 23 | 9.3456 | 8.718476 | 8.453267 | 8.198259 | 7.717264 | 7.802617 | 0 |
| 24 | 9.38743 | 8.711017 | 8.507113 | 8.132324 | 7.836855 | 7.82903 | 0 |
| 25 | 9.459962 | 9.150949 | 8.604759 | 8.086293 | 7.779938 | 7.465349 | 0 |
| 26 | 9.455664 | 9.606517 | 9.526521 | 9.177776 | 8.014919 | 7.262951 | 0 |

fingerprint_image_dataset

**Figure 2**: The Pre-processed Dataset.

The divergence values (pre-processed dataset) therefore serve as inputs into the machine learning techniques (Naive Bayes, Decision Tree, Logistic Regression and CNN) algorithms considered in this paper. The Python programming language virtual platform (Google Colaboratory) is used to implement the proposed algorithms. The goal of our proposed method is to train these machine learning techniques to carry out the intra-class separability of fingerprint images. Therefore, the input data is split into 70 % training, and 30% testing of the model. These algorithms are selected for usage because they are well suited for the labeled datasets considered in this paper. Figure 3 summarises the schematic diagram of the proposed model.

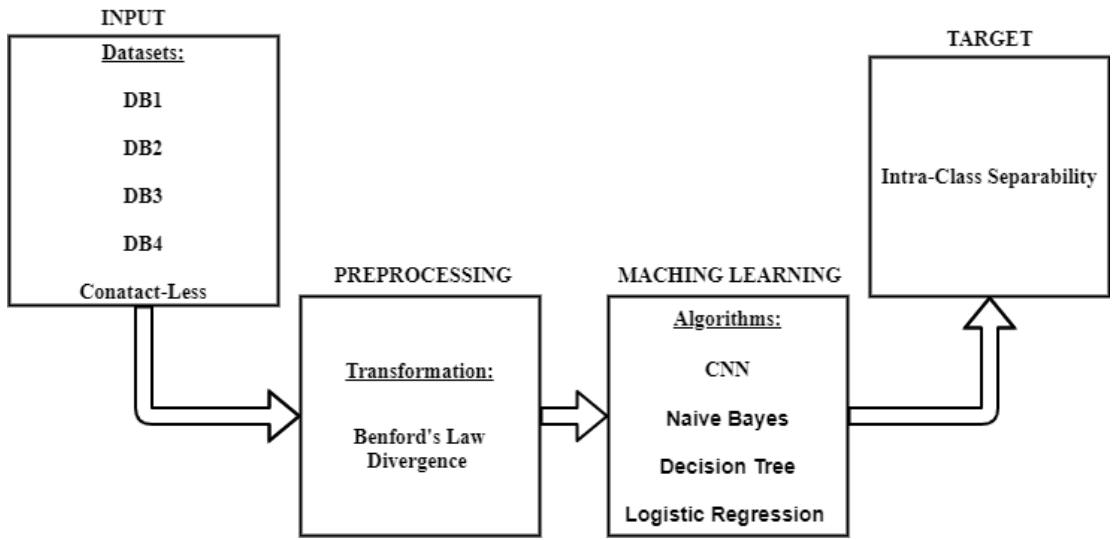

**Figure 3**: Schematic Diagram of the Proposed Model

**3.5 Evaluation Metrics**

To evaluate the proposed model, the following evaluation metrics are used:
  i. **Accuracy**: This is mathematically expressed by the formula:
      Accuracy = (TP + TN) / (TP + TN + FP +FN)

Where:

TP (True Positive): The outcome where the model correctly predicts the positive class.

TN (True Negative): The outcome where the model correctly predicts the negative class.

FP (False Positive): The outcome where the model incorrectly predicts the positive class.

FN (False Negative): The outcome where the model incorrectly predicts the negative class.

  ii. **Precision**: This is shown mathematically as:
      Precision = (TP) / (TP + FP)

iii. **Recall**: This is shown mathematically as:

Recall = (TP) / (TP + FN)

iv. **F1-Score**: This is mathematically expressed as:

F1-Score = (2 x Precision x Recall) / (Precision + Recall).

## 4. RESULTS AND DISCUSSIONS

### A. Naive Bayes Performance Results

The pre-processed fingerprint data is fed into the Naïve Bayes algorithm. As shown in Figure 4, the Naïve Bayes algorithm achieved an overall accuracy of 95.95%, with an F1 score of 0.96, the Precision value of 0.96, and a 0.96 Recall value.

```
***NAIVE BAYES RESULTS***
Accuracy: 95.95%
F1 Score: 0.958750
Precision:  0.961111111111111
Recall:  0.9609022556390977
```

**Figure 4**: Naïve Bayes Results

Again, Figure 5 shows the confusion matrix result for the Naïve Bayes algorithm. The resulting confusion matrix shows that DB1, DB2, and DB5 classes were excellently classified at an accuracy of 100 percent.

However, for DB3, 1 fingerprint image was misclassified as DB1 and 1 fingerprint image was misclassified as DB2. While in DB4, only 1 fingerprint image was misclassified as DB1.

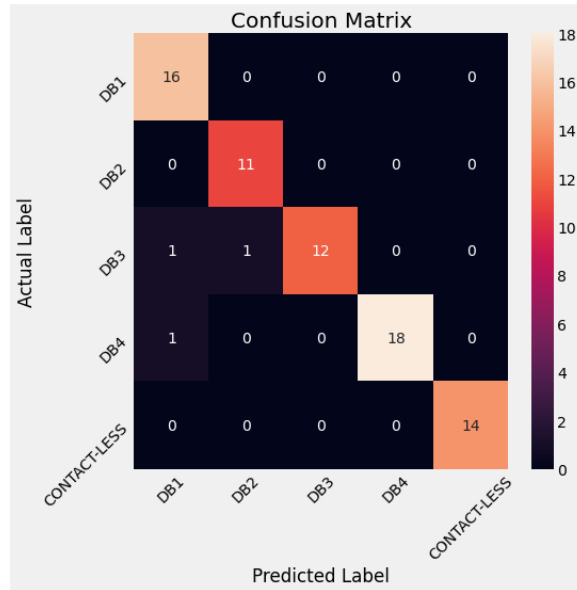

**Figure 5**: The Confusion Matrix for the Naïve Bayes Algorithm

### B.    Decision Tree Performance Results

As shown in Figure 6, the Decision Tree algorithm achieved an overall accuracy of 100.00%, with an F1 score of 1.00, the Precision value of 1.00, and a 1.00 Recall value.

```
***DECISION TREE RESULTS***
Accuracy: 100.00%
F1 Score: 1.000000
Precision:  1.0
Recall:  1.0
```

**Figure 6**: Decision Tree Results

The Decision Tree confusion matrix shows that all the dataset classes (DB1, DB2 DB3, DB4, and DB5) were excellently classified at 100% accuracy as shown in Figure 7.

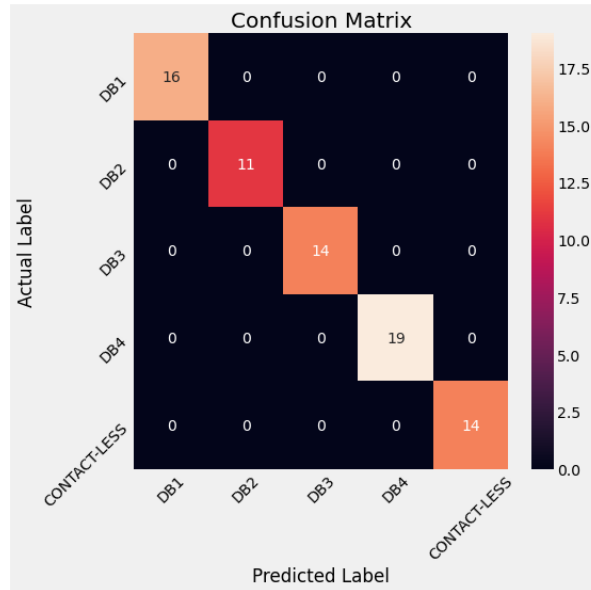

**Figure 7**: The Confusion Matrix for the Decision Tree Algorithm

### C. Logistic Regression Performance Results

The Logistic Regression algorithm achieved an overall accuracy of 90.54%, with an F1 score of 0.90, the Precision value of 0.92, and a 0.91 Recall value as shown in Figure 8.

```
***LOGISTIC REGRESSION RESULTS***
Accuracy: 90.54%
F1 Score: 0.896609
Precision:  0.9152777777777779
Recall:  0.9075187969924812
```

**Figure 8**: Logistic Regression Results

The Logistic Regression confusion matrix shows that DB1, BB2, and DB5 were accurately classified with an accuracy of 100%. For the DB3, 5 fingerprint images were misclassified as DB2 and considering DB4, 2 fingerprint images were misclassified as DB1 as shown in Figure 9.

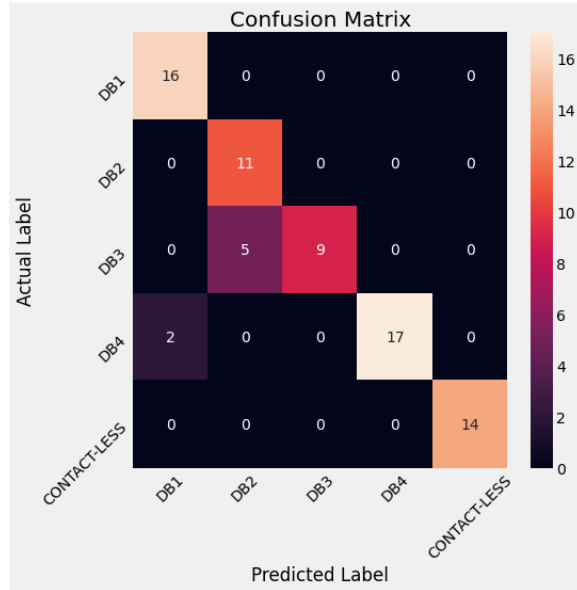

**Figure 9**: The Confusion Matrix for the Logistic Regression Algorithm

### D. Convolutional Neural Network (CNN) Results

Figure 10 shows the training loss versus epochs for the intra-class classification of fingerprint images.

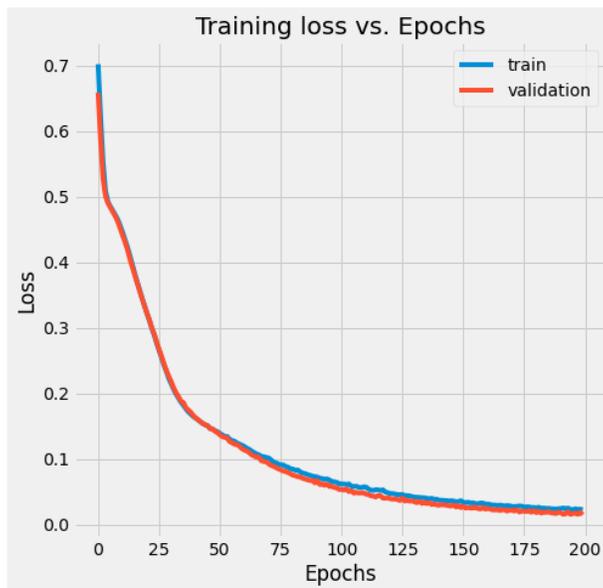

**Figure 10**: The CNN Training Loss Vs Epoch Graph

Based on the optimization capacity of the CNN, we always expect a lower loss to produce a better model, especially when considering the training vs. epochs.

We can see in Figure 10 that the loss tends more towards zero around 100 epochs and more. This shows that our proposed model performed well for the training and validation datasets, especially after 100 epochs.

Furthermore, the training accuracy vs. epochs considered in this experiment, for both the training and validation datasets, shows that epochs above 75 show more consistent and higher results that are closer to 1, as shown in Figure 11.

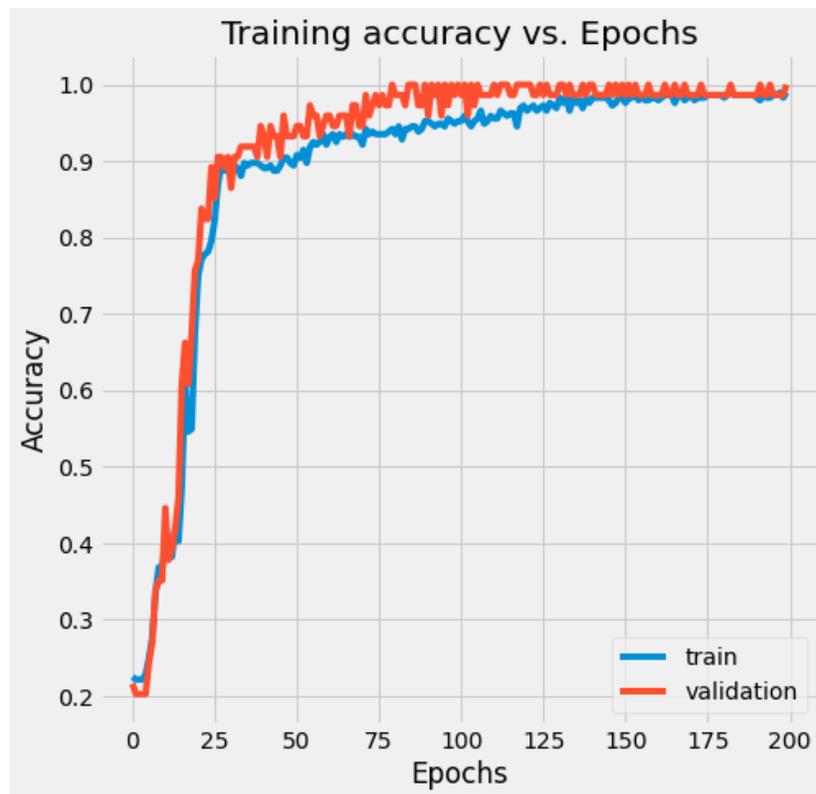

**Figure 11**: The CNN Training Loss vs Epochs Graph

Again, Figure 12 shows the confusion matrix for the intra-class classification using CNN.

The confusion matrix shows that all the 80 fingerprint images in DB1, DB2, DB4, and DB5 are correctly classified at 100% accuracy. While the DB3 fingerprint images were correctly classified at 94% accuracy and 6% of the DB3 fingerprint images were misclassified as DB2. The CNN confusion matrix has proved that the novel proposed method indeed performed the intra-class classification accurately.

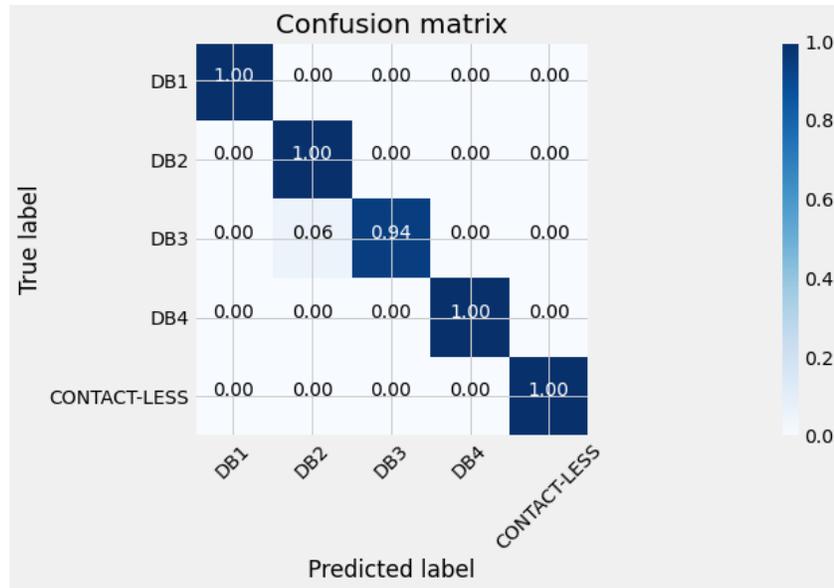

**Figure 12**: The Confusion Matrix for the CNN Algorithm

From the experiments as shown in the confusion matrix in Figure 12, Benford's law divergence values and CNN excellently classified the fingerprint images. From these results, we observed that the CNN algorithm and the Decision Tree algorithm outperformed the other algorithms having achieved an overall accuracy of 100%.

## 5. Conclusion and Future Work

This paper proposed a novel use of Benford's law divergence values and their application for classifying fingerprint images by characteristics and/or sensor sources using Naive Bayes, Decision Tree, Logistic Regression, and CNN algorithms. It was shown that the Naive Bayes, Decision Tree, Logistic Regression and CNN algorithms successfully classified the fingerprint images with an accuracy of 95.95%, 100.00%, 90.54%, and 100%, respectively.

This shows that our proposed method can effectively reduce features and achieve high intra-class separability results especially using Decision Tree and CNN algorithms. For future work, we plan to investigate other classification techniques such as Long Short-Term Memory (LSTM) for the intra-class classification and source identification of fingerprint images.